\documentclass[runningheads]{llncs}
\usepackage{graphicx}
\usepackage{amsmath,amssymb} 
\usepackage{color}
\usepackage{multirow}
\usepackage{float}
\usepackage[width=122mm,left=12mm,paperwidth=146mm,height=193mm,top=12mm,paperheight=217mm]{geometry}
\newcommand{\specialcell}[2][c]{%
\begin{tabular}[#1]{@{}l@{}}#2\end{tabular}
}	
\begin{document}
\pagestyle{headings}
\mainmatter

\title{SFace: An Efficient Network for Face Detection in Large Scale Variations} 

\titlerunning{A very long title}

\authorrunning{authors running}




\author{
Jianfeng Wang$^1$$^2$$^*$,  
Ye Yuan $^1$$^\dagger$,
Boxun Li$^\dagger$, 
Gang Yu$^\dagger$ 
\and Sun Jian$^\dagger$\\
}

\institute{
College of Software, Beihang University$^*$\\
Megvii Inc. (Face++)$^\dagger$\\
\email{ \{wjfwzzc\}@buaa.edu.cn},
\email{ \{yuanye,liboxun,yugang,sunjian\}@megvii.com}
}
\footnotetext[1]{Equal contribution.}
\footnotetext[2]{Work was done during an internship at Megvii Research.}

\maketitle

\begin{abstract}
Face detection serves as a fundamental research topic for many applications like face recognition. Impressive progress has been made especially with the recent development of convolutional neural networks. However, the issue of large scale variations, which widely exists in high resolution images/videos, has not been well addressed in the literature. In this paper, we present a novel algorithm called SFace, which efficiently integrates the anchor-based method and anchor-free method to address the scale issues. A new dataset called 4K-Face is also introduced to evaluate the performance of face detection with extreme large scale variations. The SFace architecture shows promising results on the new 4K-Face benchmarks. In addition, our method can run at $\sim$50 frames per second (fps) with an accuracy of 80\% AP on the standard WIDER FACE dataset, which outperforms the state-of-art algorithms by almost one order of magnitude in speed while achieves comparative performance.

\end{abstract}

\begin{figure}[t]
\begin{center}
    \includegraphics[width=0.7\linewidth]{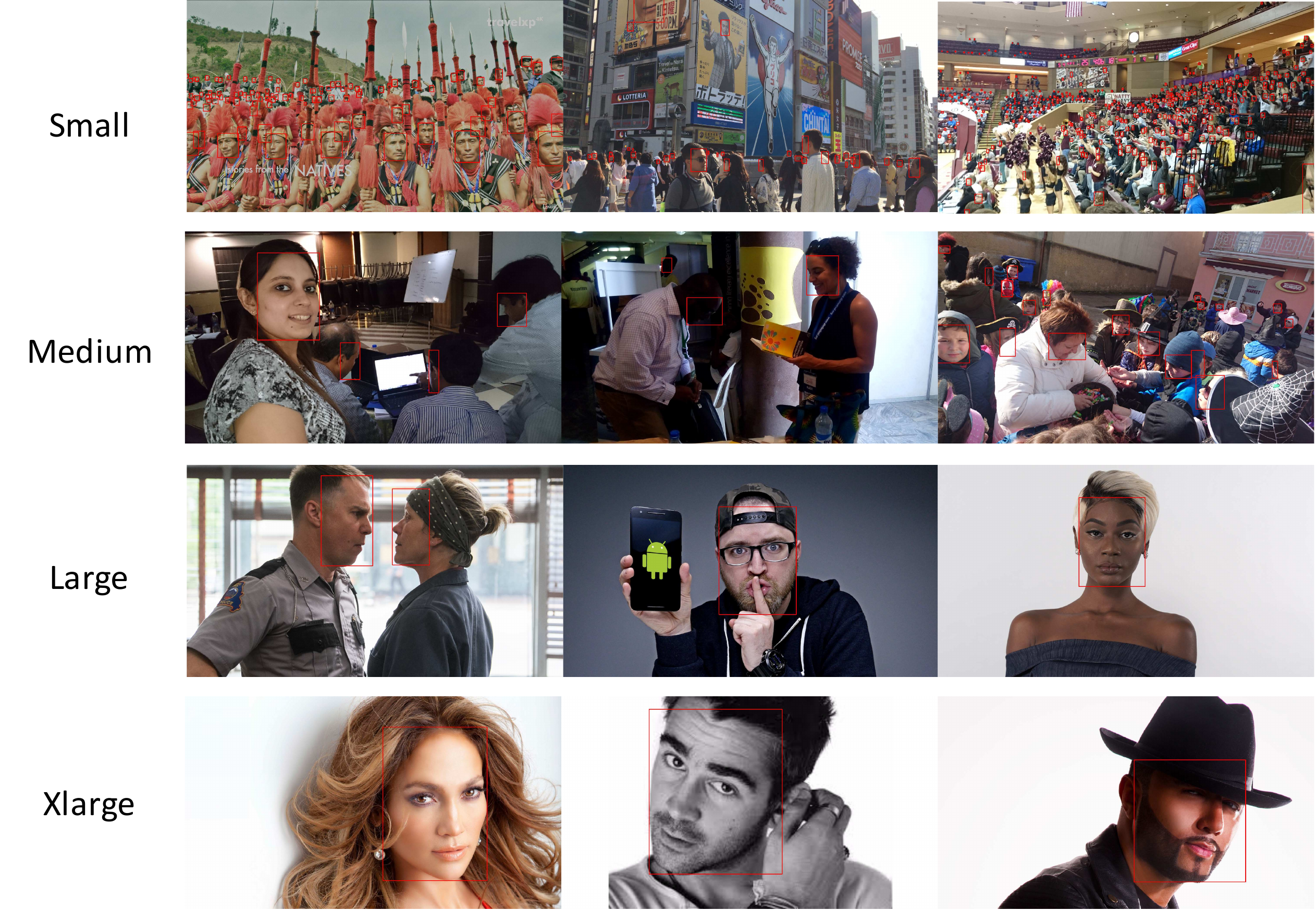}
\end{center} 
    \caption{Illustrative examples of face detection in large scale variations. All the images are from the new 4K-Face dataset.}
\label{fig:dataset_}
\end{figure}

\section{Introduction}

With the recent development of digital camera industry, the 4K ultra HD resolution cameras, e.g., 3840$\times$2160, is becoming more and more popular. These high-resolution images propose a great challenge to the face detection problem as the scales of face can range from 10$\times$10 to 2000$\times$2000. However,
the large scale variation is still one of the core challenges for face detection and has not been well solved in literature.

Traditionally, there are potentially two ways to address the scale variation problem. From the input-level, image pyramids can be applied to deal with different scales of face by using different input image sizes. However, these methods will significantly increase the computational cost and have lower inference speed. Furthermore, a sophisticated post-processing step is also needed to merge the results from different image pyramids. 

On the other hand, many works focuses on solving the scale challenge from the model (or feature) level. These methods either design specific settings of anchors or leverage anchor-free architecture to process different scales implicitly. The anchors in these algorithms must be designed carefully for each specific task, and usually can not be transferable to other datasets. This makes the anchor-based methods vulnerable to dataset distributions. On the contrary, the anchor-free algorithms, such as DenseBox~\cite{huang2015densebox} and UnitBox~\cite{yu2016unitbox}, usually fail to obtain the accurate localization ability compared with anchor-based methods. As a results, the anchor-free methods have trailed the accuracy of anchor-based detectors thus far.

In this paper, we present a novel algorithm called SFace to address the scale variation issue efficiently. More specifically, the SFace architecture integrates anchor-based methods (like RetinaNet~\cite{lin2017focal}) and anchor-free based methods (like UnitBox~\cite{yu2016unitbox}) with two branches. The idea is inspired by the following observation. The anchor-based methods can provide accurate bounding-box localization for the face scales ranging from 32$\times$32 to 512$\times$512 with a common anchor setting. And the faces with arbitrary sizes, especially for the faces with extreme large scales, can be implicitly captured by anchor-free methods. By combining the two methods efficiently, our method achieves high detection performance while maintains low computational cost as well. For better merging the two branches, we also present an effective re-score approach based on the Intersection-Over-Union (IOU) prediction. The proposed re-score approach can efficiently unify the confidence scores of two different branches and leads to a both better and easier merging process. 

In addition, systematic benchmarks with large scale variations for face detection in high resolution images are still lacking. Therefore, we also present a new benchmark, called 4K-Face, to evaluate face detectors in extreme large scale variations. The 4K-Face dataset is annotated with the WIDER FACE style. The dataset includes around 5000 ultra high resolution images with extremely large face scale variations. To the best of our knowledge, this is the first dataset designed for 4K high resolution face detection.

The main contributions of this paper can be summarized as follows.
\begin{itemize}
    \item We present a novel architecture, called SFace, to address the large scale variations by efficiently integrating anchor-based method and anchor-free method. An effective re-score method based on the Intersection-Over-Union (IOU) prediction is proposed on top of the model to better unify the outputs of two branches. 
    \item A new benchmark, named 4K-Face, with around 5,000 images and 30,000 face annotations. This is the first benchmark that aims to explicitly evaluate face detectors in high resolution images with extremely large scale variation of faces.
    \item Promising results have been reported with fast inference speed. Our method obtains the AP of $\sim$80\% at the speed of 50fps on the public WIDER FACE benchmark~\cite{yang2016wider}. 
\end{itemize}

\section{Related Work}

Face detection is a fundamental and essential step for many face related applications, e.g. face landmark~\cite{xiong2013supervised,zhu2016face} and face recognition~\cite{parkhi2015deep,schroff2015facenet,zhu2015high}. The milestone work of Viola-Jones~\cite{viola2001rapid} utilizes AdaBoost with Haar feature to get a real-time face detector with good accuracy. After that, lots of works have been proposed to improve the performance by introducing more powerful classifiers and more sophisticated hand-crafted features~\cite{brubaker2008design,zhu2006fast,liao2016fast}. Besides, \cite{felzenszwalb2008discriminatively} employed deformable part models (DPM) in face detection and achieved remarkable performance. Compared to previous hand-crafted features, the CNN-based algorithms have demonstrated great boost of both accuracy and  robustness in face detection. These methods may be summarized according to the following categories: the anchor-based methods, and the anchor-free algorithms.

\subsubsection{Anchor-based methods.} A series of works, such as Faster RCNN~\cite{ren2015faster}, SSD~\cite{liu2016ssd}, DSSD~\cite{fu2017dssd}, FPN~\cite{lin2017feature} and RetinaNet~\cite{lin2017focal}, use pre-designed region proposals, which are called anchors, to predict the location of object targets. In these methods, the network will be trained to regress the offsets between the anchors and ground truth bounding boxes. State-of-the-art results have been reported by the anchor-based methods. However, the settings of anchors must be designed carefully for each specific task to achieve good performance. Moreover, since the scales and aspect ratios of anchors are fixed, it is difficult for these anchor-based methods to handle object candidates with large shape variations, especially for small objects like faces~\cite{yu2016unitbox}.

\subsubsection{Anchor-free methods.} In contrast, several algorithms, such as YOLO~\cite{redmon2016you}, YOLO9000~\cite{redmon2017yolo9000}, DenseBox~\cite{huang2015densebox}, and UnitBox~\cite{yu2016unitbox}, directly output the bounding boxes without anchors. There is no restriction of pre-designed scales or ratios in these methods. For example, UnitBox~\cite{yu2016unitbox} utilizes every pixel of the feature map to regress a 4-D distance vector, which represents the distances between the current pixel and the four bounds, to localize the object candidate containing the pixel. The anchor-free methods usually have better speed performance by removing the dense anchors. However, it is also more difficult for the network to learn to regress the huge variations of bounding boxes without any prior knowledge. As a result, the detection performance of anchor-free methods, especially for the localization accuracy, is still slightly suppressed by the anchor-based method.

\textbf{\begin{figure}[tb]
\begin{center}
    \includegraphics[width=0.9\linewidth]{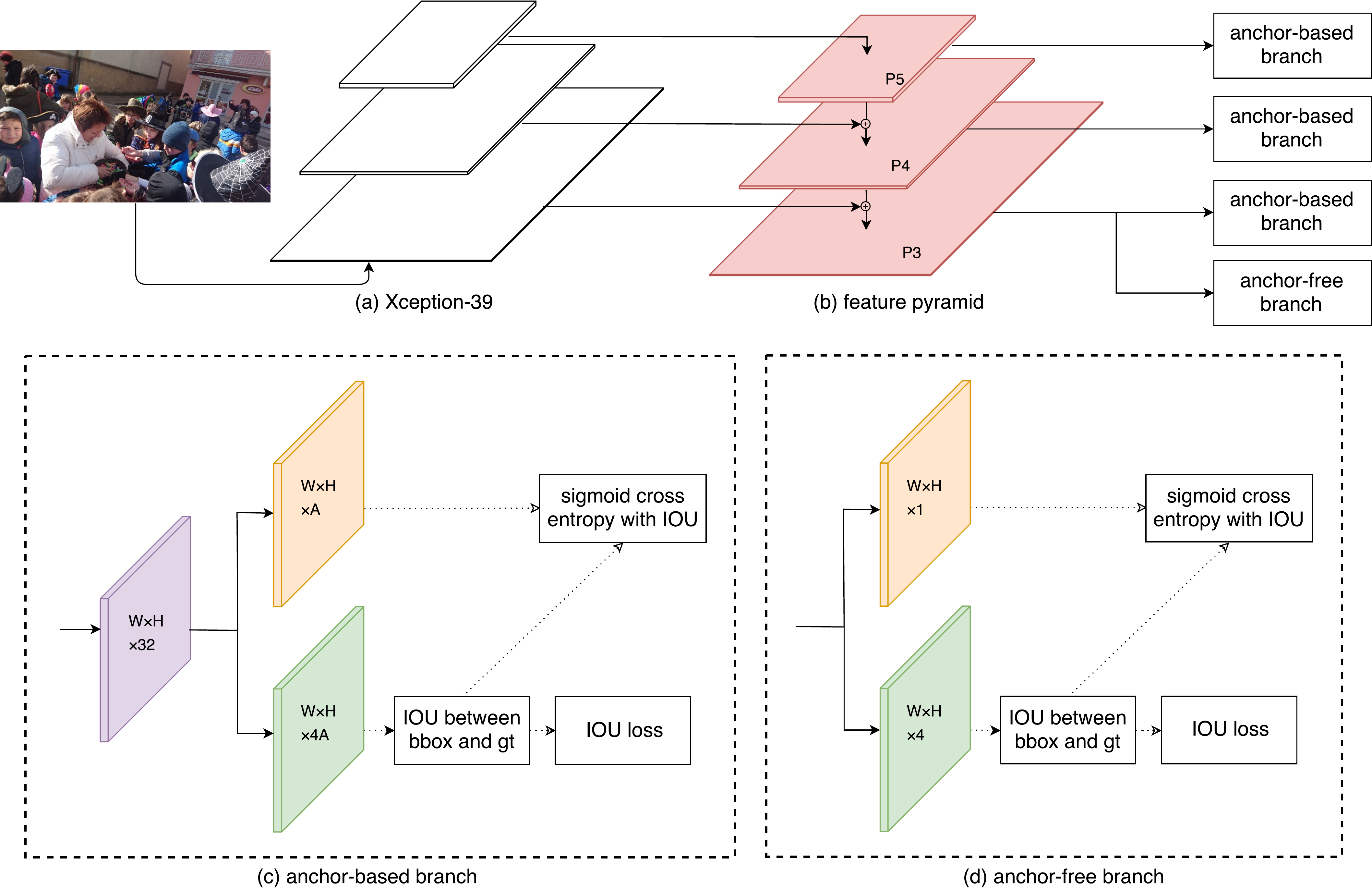}
\end{center} 
    \caption{An overview of our SFace network architecture.}
\label{fig:network_}
\end{figure}
}

\subsubsection{Handling scale variance in face detection.}  The  issue  of  large  scale  variations  of  faces has not been well addressed in literature. A lot of recent approaches focus on solving the scale variation problem. On the one hand, it is intuitive to detect faces in different scales by leveraging image pyramids. previous works, such as S3FD~\cite{zhang2017s3fd} and SSH~\cite{najibi2017ssh}, reported considerable performance improvement in wide scaling datasets.  However, the multi-scale test will bring huge computational cost and lead to lower inference speed. On the other hand, other works focuses on learning stronger scale invariant features. Qin~\cite{zhang2016joint} proposed a joint cascade network for learning multi-scale features. Objects with different scales are handled by different subnet, separately. Besides, Lin et al.~\cite{lin2017feature} proposed feature pyramid network to better fuse multi-layer features and reported remarkable object detection performance. Nevertheless, there methods still relay on a dense and careful anchor setting. The performance would plummet if the object is out of the range of pre-designed anchors.


\section{SFace Architecture}

Our goal is to achieve the accuracy of anchor-based method without careful attention to the anchor setting. To be specific, we present a novel algorithm called SFace, which integrates anchor-based  methods  (like RetinaNet~\cite{lin2017focal})  and  anchor-free  based  methods  (like DenseBox~\cite{huang2015densebox})  in  two  branches. The construction of the proposed architecture involves an FPN backbone, a set of anchor based branches, and one anchor-free branch.


\subsection{Feature Pyramid Backbone\label{sec:fpn backbone}} 

Convolutional neural networks extract different levels of features and spatial resolutions in different network layers. The bottom layers of the network maintain high spatial resolution, which is good for spatial localization, especially for the small objects. On the contrary, the top layers of the network can extract powerful semantic features, which has demonstrated great power for classification tasks. previous work~\cite{lin2017feature} proposed an effective backbone, called feature pyramid network (FPN), to fuse features from different levels. FPN creates a powerful feature pyramid that has strong semantics at all scales.

Following the design principle of FPN, our method uses a top-down architecture with lateral connections to build an in-network feature pyramid from a single-scale input. Specifically, we create FPN from P3 to P5 (from stride 8 to stride 32) and introduce high level semantic features (Conv5) to the bottom layer (Conv3) with high resolution. Different from previous works like RetinaNet, the proposed method does not create branches on top of P5 or extract features at very high resolution. Our experiments show that these layers (P3-P5) are sufficient to detect most of faces with good efficiency and are robust to dataset distribution. Meanwhile, in order to further reduce the computational cost, we only use a $1 \times 1$ convolutions with 32-dimension output channels as lateral connections of the feature pyramid blocks. The summation of the upsample of P$_n$, and the 32-D convolutions from Conv$_n$, will be used as the fused feature pyramids, and then directly feeds the classification or regression subnets.

Besides the feature pyramid network, we use an Xception-like base model~\cite{chollet2017xception}, namely the Xception-39 \footnote{The network is called Xception-39 because its computational flops is 39M.}, to achieve good balance between speed and accuracy. The architecture and specifics of the network are illustrated in Table~\ref{table:xception-39}. The model has deep layers with large receptive field of 1679. It achieves a Top-1 error of 44.9\%, and Top-5 error of 21.4\%, under single crop on the ImageNet validation dataset~\cite{russakovsky2015imagenet}.


\setlength{\tabcolsep}{8pt}
\begin{table}[th]
\centering

\begin{tabular}{lcccccccc}
\hline
Layer & Output size & KSize & Stride & Repeat & \specialcell{output\\channels} \\
\hline
Image & $224\times 224$ & & &  \\
\hline

Conv1 & $112\times 112$ & $3\times 3$ & 2 & 1 & 8 & \\
MaxPool & $56\times 56$ & $3\times 3$ & 2 &  &  \\
\hline
P3 & $28\times 28$ & & 2 & 1 & 16  \\
 & $28\times 28$ & & 1 & 3 & 16 \\
\hline
P4 & $14\times 14$ & & 2 & 1 & 32 &  \\
 & $14\times 14$ & & 1 & 7 & 32 &  \\
\hline
P5 & $7\times 7$ & & 2 & 1 & 64  \\
& $7\times 7$ & & 1 & 3 & 64 \\
\hline
GAP & $1\times1$ & $7\times 7$ & & &  \\ 
\hline
FC & & & & & 1000  \\
\hline
\hline
FLOPs & & & & & 39M \\
\hline
\end{tabular}
\caption{Architecture description of the Xception-39 backbone network.}
\label{table:xception-39}
\end{table}

\subsection{Anchor-based Branches\label{sec:anchor-based branch}}

We employ an anchor branch to get better localization performance. Our goal is to leverage the accuracy of anchor-based method without careful attention to the anchor setting. Therefore, the anchor branch is largely reduced compared with previous methods~\cite{lin2017focal,zhang2017s3fd,hao2017scale,girshick2015fast}.

To be specific, our method outputs anchor-based branches on feature pyramids from P3 to P5. The redundant layers for large faces, such as P6 and P7 (with stride 64 and 128, respectively) are removed. The anchors have areas of $16^2$ to $64^2$ from P3 to P5. Each pyramid level have anchors of 2 ratios \{1:1, 1:1.5\} and 3 scales \{$2^{0}$, $2^{1/3}$, $2^{2/3}$\}. In summary, the anchor setting requires 6 anchors per pyramid level and is able to capture face areas from $16^2$ to $101.59^2$ pixels with respect to the network input.

During training, each anchor will be assigned a binary classification target (background or face) and a 4-D vector of bounding box regression target. We adjust the assignment rule of previous anchor-based methods for better merging result. Moreover, we also deploy IOU Loss, instead of the original Smooth L1 Loss, in the bounding box regression subnet. More details will be discussed in Section~\ref{sec:classification with IOU}.



\subsection{Anchor-free Branch\label{sec:anchor-free branch}}

An anchor-free branch is equipped with feature pyramid level P3 to assist the branches with anchors. This anchor-free branch is used to capture objects whose scales can not be covered by pre-designed anchors, especially the faces that have a very large scale. 

Inspired by previous work~\cite{huang2015densebox,yu2016unitbox}, the anchor-free branch on P3 directly regresses a 4-D vector that represents the distances between the current pixel and the four bounds of object target. For example, a target bounding box in output coordinate space can be represented with the left-top pixel $(x_t, y_t)$ and the bottom-right pixel $(x_b, y_b)$. Then the pixel, located in the corresponding areas of the target bounding with the coordinate of $(x_i, y_i)$ in the output feature map, can describe the target bounding box with a 4-dimensional vector $\hat{t}$:
\begin{equation}
    \hat{t} = \{d_{xt} = x_i - x_t, d_{yt} = y_i - y_t, d_{xb} = x_b - x_i, d_{yb} = y_b - y_i\}
\end{equation}  

We employ IOU Loss for distance regression of $\hat{t}$ to balance bounding boxes with varied scales. The IOU Loss can be described as follows:
\begin{equation}
L_{IOU}=-\frac{1}{N}\sum{\ln{\frac{\rm{intersection(pred, target)}}{\rm{union(pred, target)}}}}
\end{equation}
The IOU Loss will normalize the loss of boxes with different scales by their areas, and show robust to
objects of varied shapes and scales. More details will be described in the following Section~\ref{sec:classification with IOU}.


\subsection{Classification with IOU\label{sec:classification with IOU}}

The outputs of anchor-based branches and anchor-free branches have significant difference in both localization manners and confidence scores, which leads to great difficulty of merging two outputs together. The reasons are as follows:
\begin{itemize}
    \item For the original anchor-based methods, the anchors that fall in the IOU of [0.5, 1] will be regarded as positive training samples. The definition of positive and negative samples has no relation to the results of bounding box regression. As a result, the classification confidence of the anchor-based branch mainly represents the probability of whether the corresponding anchors, not the final prediction, successfully catch faces. It is hard to evaluate the final localization accuracy with the classification confidence score. The situation is even more serious for networks that separate the classification and regression subnets earlier.  
     \item For the anchor-free methods like DenseBox and UnitBox, the network is trained in a segmentation-like way. The positive labeled region in the output feature map is defined to a filled ellipse with radius located in the center of a face bounding box. For example, for a given bounding box, the positive `segmap' of the face is an ellipse lies in the center of the bounding box. The radius of the positive ellipse is proportional to the box scale (e.g., 0.3 to the box width and height \cite{huang2015densebox}). The other pixels will be treated as background. In this way, the confidence score of the anchor-free branch mostly indicates whether the corresponding pixel falls on a face, and also has weak relationship to the localization prediction.
\end{itemize}

To summarize, neither the anchor-based method, nor the anchor-free method, includes the final localization accuracy in the classification subnet. And their confidence scores represent different information. Directly merging the bounding boxes of the two branches by their classification results is unreasonable and may leads to drastically drop of detection performance.

Therefore, we adjust the classification subnets of the two branches to regress the IOU scores between the prediction and target. Specifically, both the anchor-based branch and the anchor-free branch will generate a bounding box prediction in the first step of training\footnote{The anchor-based branch will first filter anchors with an IOU threshold of [0.5, 1]. The kept anchors will be used to calculate the final IOU prediction score}. We then calculate their IOUs between the ground-truth targets. The anchors (in the anchor-based branch) and the pixels (in the anchor-free branch), whose IOU is larger than 0.5, will be regarded as the positive sample and the others are negative. A cross-entropy loss is deployed to discriminate positive and negative samples as binary classes\footnote{We tried to use IOU scores directly as regression targets for the classification subnet. The classification subnet will be trained in a regression manner. However, experimental results demonstrate performance drop compared to the binary cross-entropy loss.}.

We use IOU Loss for bounding box regression in both anchor-based branches and anchor-free methods. This adjustment helps unify the output manner of the two branches and make better combination results. Following the design principle of previous works, the bounding box regression loss is only defined on the positive samples. The other anchors or pixels, whose IOUs fall in [0, 0.5), have no contribute to the bounding box regression loss of the network. We also employ focal loss~\cite{lin2017focal} to overcome the extreme foreground-background class imbalance encountered in training.


\section{Experiments}

The proposed method is evaluated on two benchmarks: WIDER FACE, and 4K-Face. The WIDER FACE is widely used in face detection. And 4K-Face is a new dataset which is specifically designed to evaluate face detectors for faces with large scale variations.
\subsection{Datasets}

\subsubsection{WIDER FACE}

The WIDER FACE dataset~\cite{yang2016wider} contains 32,203 images and 393,703 annotated faces with a high degree of variation in scale, pose and occlusion. 158,989 of the images are chosen as the training set, 39,496 images are in the validation set, and the rest are used for testing. The validation set and test set are split into `easy’, `medium’ and `hard’ subsets cumulatively (the `hard’ set contains all images). The WIDER FACE is one of the most challenging face datasets for face detection.

\subsubsection{4K-Face: A Dataset with Huge Scale-varying Faces}

\textbf{\begin{figure}
\begin{center}
    \includegraphics[width=0.7\linewidth]{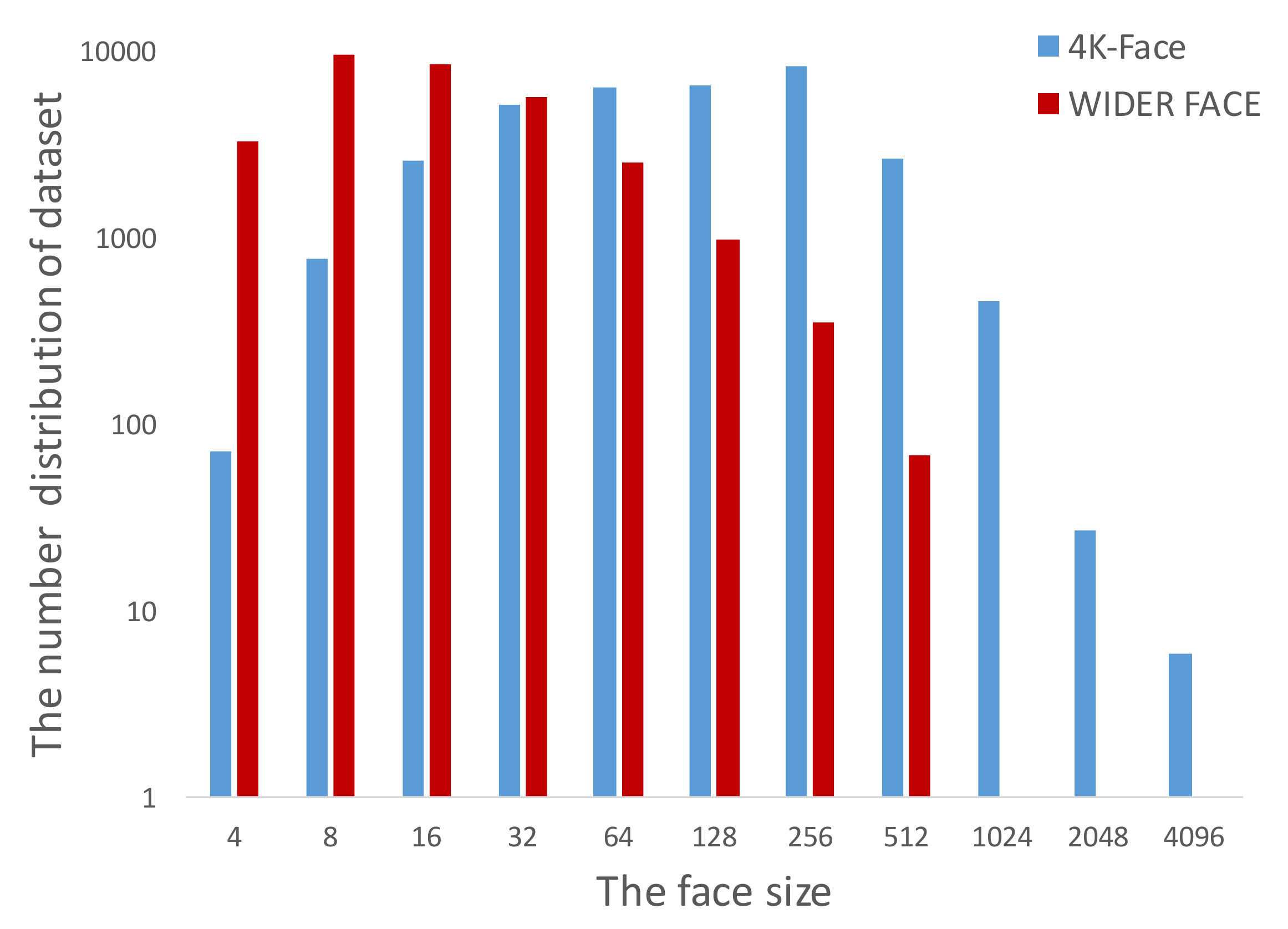}
\end{center} 
    \caption{The distribution of face size in WIDER FACE val dataset and 4K-Face. The X axis represents the face size, and the Y axis represents the number of corresponding faces in the dataset. The Y axis uses logarithmic scale.}
\label{fig:dataset_dist}
\end{figure}
}
We observe that there is a gap between current face detection benchmark and the real world images, especially for high resolution images and videos. Thus, few datasets are targeted for faces with huge scale variation in literature. As illustrated in Figure~\ref{fig:dataset_dist}, only $\sim$1\% annotated faces in the WIDER FACE dataset are larger than 512 pixels. In contrast, more than 30\% face boxes are smaller than 32 pixels. The scale distribution is very unbalanced. To solve this problem, we introduce a new dataset, called 4K-Face, for benchmarking the face detector with huge scale variance.

\subsubsection{Data Collection of 4K-Face}
The images from 4K-Face is collected from the Internet. Following WIDER FACE event categories, keywords, such as travel, surgeons, and celebration, are used to retrieve more than 25,000 images from web search engine. In order to obtain large scale variance, we only keep the images with 4K resolution (3840$\times$2160). Images without face are filtered. Finally, 5,102 images with more than 30,000 annotated boxes are acquired in total.

Comparing with WIDER FACE, the 4K-Face dataset is more challenging in terms of scale variation. As shown in Figure~\ref{fig:dataset_dist}, the face scale distribution in the 4K-Face is more balanced. More face with large scale (larger than 512 pixels) exist in the dataset. These factors are essential to simulate and satisfy the requirements of real world system.

It should be noted that our method is trained on the WIDER FACE training set. The 4K-Face dataset is only used for evaluation and ablation study.

\subsection{Implement Details}

We use Xception-39 model pre-trained on ImageNet dataset as the backbone network. The architecture of the model is illustrated in Table~\ref{table:xception-39}. The Top-1 and Top-5 error rates are 44.9\% and 21.4\%, respectively, on  the  ImageNet validation dataset. All the detection networks are trained with AdamW~\cite{loshchilov2017fixing} optimizer on 4 GPUs with 128 images per mini-batch (32 images per GPU). The learning rate is initialized to $1e-3$, and is dropped only once by 10 after 60k iterations. The weight decay is set to $1e-5$. Similar to RetinaNet, the weights in the outputs of both classification and regression subnets are initialized with bias $b = 0$ and a Gaussian distribution with variance $\sigma = 0.01$. 

Our network is trained on the training set of WIDER FACE with the following data augmentation strategy:
\begin{itemize}
    \item We random crop a square patch from the original image, and only keep the ground-truth boxes whose centers are inside the selected patch. Then we resize the patch to $600\times600$ for training.
    \item We horizontally flip the patch with a probability of 0.5.
    \item We apply color jitter described in \cite{howard2013some}. Specifically, we disturb the contrast, the brightness, and the color in a random order. Each property is multiplied by a random coefficient in the range of $[0.5, 1.5]$ independently.
\end{itemize}

For testing, we perform standard Non-Maximum Suppression (NMS)~\cite{neubeck2006efficient} with an IOU threshold of 0.5 for merging outputs from different branches. All models are tested on WIDER FACE images with shortest edge resize to 1500 pixels, and 4K-Face images with original sizes.

\begin{table}[t]
\begin{center}
\begin{tabular}{l|c|c|c}
\hline\hline
Method & AP(easy) & AP (medium) & AP(hard) \\
\hline\hline
ACF~\cite{yang2014aggregate} & 69.5 & 58.8 & 29.0 \\
Faceness~\cite{yang2015facial} & 71.6 & 60.4 & 31.5 \\
LDCF+~\cite{ohn2016boost} & 79.7 & 77.2 & 56.4 \\
MT-CNN~\cite{zhang2016joint} & 85.1 & 82.0 & 60.7 \\
CMS-RCNN~\cite{zhu2017cms} & 90.2 & 87.4 & 64.3 \\
ScaleFaces~\cite{yang2017face} & 86.7 & 86.6 & 76.4 \\
\hline
SFace & \textbf{89.1} & \textbf{87.9} & \textbf{80.4} \\
\hline
\end{tabular}
\end{center}
\caption{Comparison of SCaleFace with state-of-art detectors on the test set of the WIDER FACE dataset.}
\label{table:WIDER FACE_result}
\end{table}

\subsection{Overall performance}
\begin{figure*}[tp]
\begin{center}
    \includegraphics[width=0.95\linewidth]{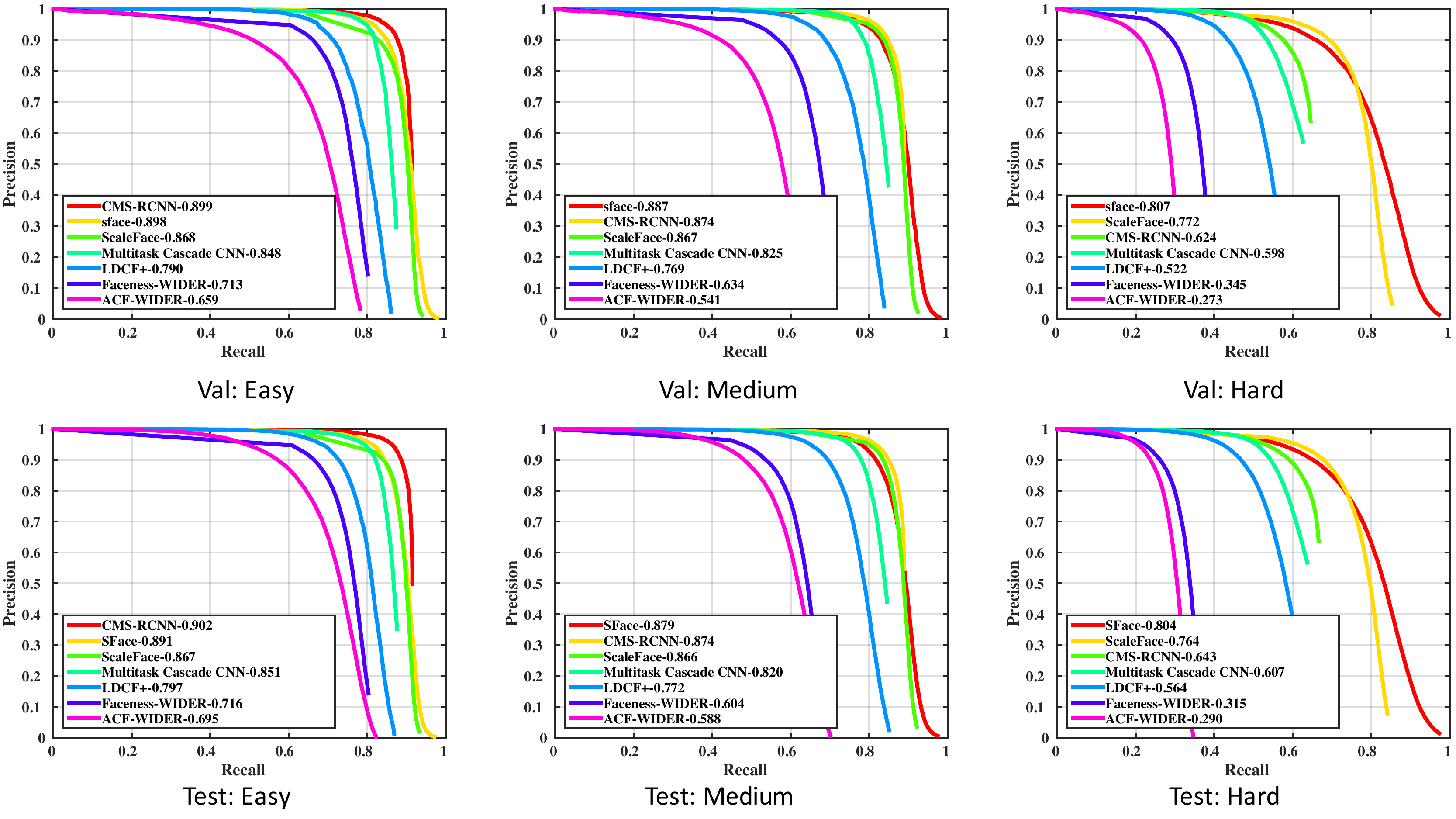}
\end{center} 
    \caption{Precision-recall curves on WIDER FACE validation and test sets.}
\label{fig:WIDER FACE_pr}
\end{figure*}

\begin{figure*}[b]
\begin{center}
    \includegraphics[width=0.9\linewidth]{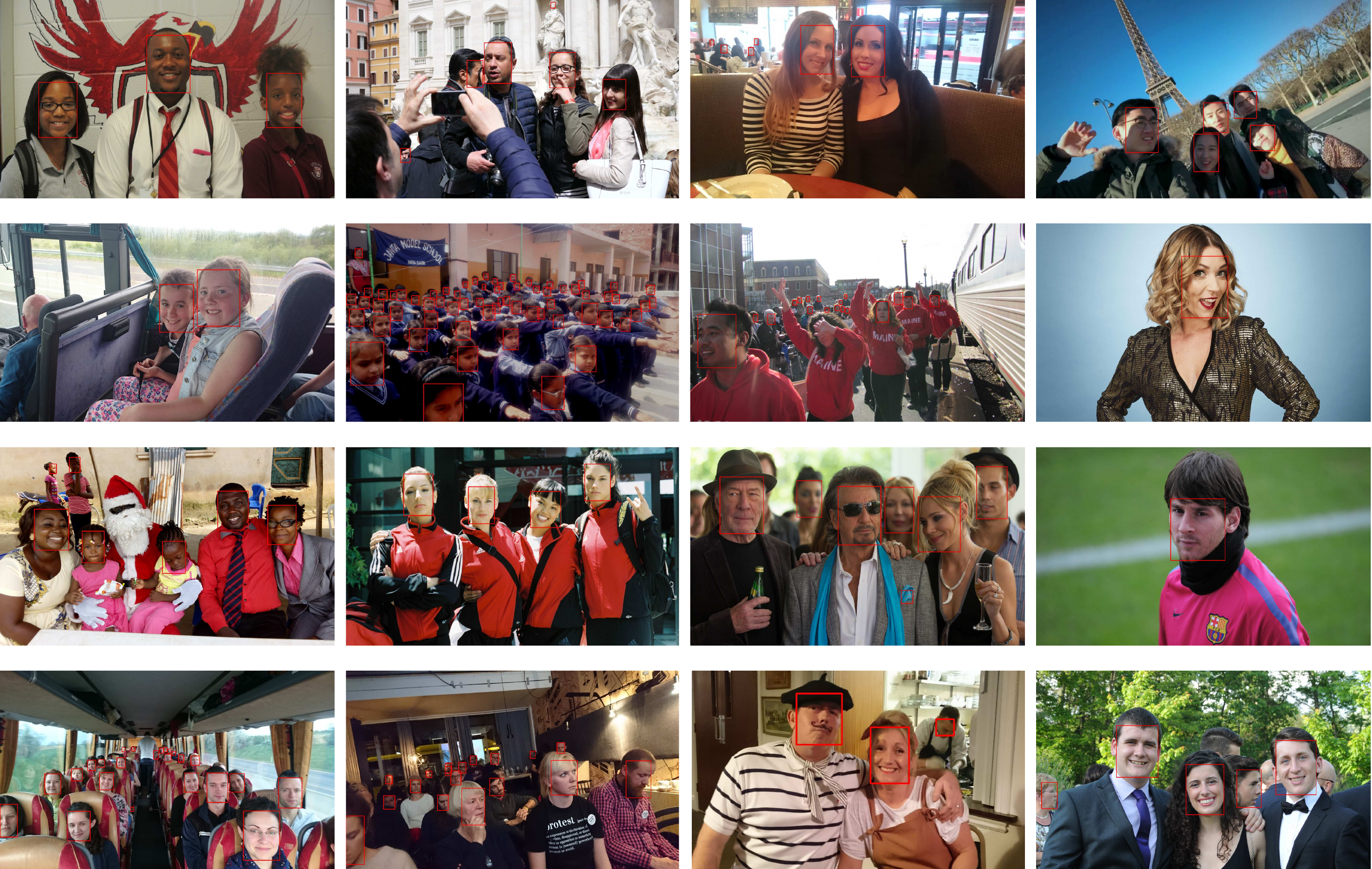}
\end{center} 
    \caption{Example result of SFace on the 4K-Face dataset.}
\label{fig:4K-Face_}
\end{figure*}

The precision-recall curves of WIDER FACE dataset is shown in Figure~\ref{fig:WIDER FACE_pr} and Table~\ref{table:WIDER FACE_result}. Our method acquires a comparable AP of 80.7\%, and significantly outperforms other methods in run-time speed. It should be noted that the backbone model of our method (the Xception-39) is more than 200 times smaller in terms of computation cost than other methods~\cite{Hu_2017_CVPR,najibi2017ssh,zhang2017s3fd}, which use base models like ResNet101~\cite{he2016deep} and VGG16~\cite{yang2014aggregate}\footnote{The computational flops of ResNet101 and VGG16 are 7.6G and 15.3G separately. In contrast, Xception-39 only has 39M Flops of computation.}. 
Meanwhile, our algorithm reaches an AP of 65.4\% on the 4K-Face dataset, suppressing the RetinaNet and Unitbox by $\sim$12\% and $\sim$2\%, respectively. Example results of our method are illustrated in Figure~\ref{fig:4K-Face_}.

\subsection{Ablation study}

\subsubsection{Compare with RetinaNet \& UnitBox}
To better understand the contribution of anchor-based branches and anchor-free branches, we compare our method with RetinaNet~\cite{lin2017focal} and UnitBox~\cite{yu2016unitbox}. For the sake of fairness, we use the same FPN-backbone setup as mentioned in Section~\ref{sec:fpn backbone}, and keep the other parts unchanged in each method. 

The \textbf{RetinaNet origin} model follows the anchor assign strategy and anchor setting described in \cite{lin2017focal}. In contrast, we also employ a \textbf{modified RetinaNet} model with similar anchor setting as describe in Section~\ref{sec:anchor-based branch}. Specifically, we only keep pyramid layers of P3, P4, and P5, and make corresponding modifications to the anchor scales. For the \textbf{UnitBox}, we use the same setting in \cite{yu2016unitbox}. All of these models are trained on WIDER FACE training set.

The results are illustrated in Table~\ref{table:tech_path_res} and~\ref{table:tech_path_4k}.

We first compare the detectors’ performance in different scales. The RetinaNet demonstrates good performance when the scales can be captured by the pre-designed anchors. However, the performance of RetinaNet drops drastically once the face scales fall out of the range of anchors. For example, RetinaNet can obtain 92.5\% and 91.2\% in easy and medium set of WIDER FACE, but only achieve 65.0\% in the hard set. In contrast, the UnitBox~\cite{yu2016unitbox} is able to detect faces with a large scale variation. But the performance in terms of AP is a little worse
for its poor localization ability compared with RetinaNet. Only AP of 70.6\% and 76.0\% are achieved in the easy and medium set, respectively. 

Compared with the two method, our method can tolerate huge scale variation and acquire a accurate localization results simultaneously. With an efficient integration of anchor-based method and anchor-free method, the proposed method acquire an AP of 80.7\% in the WIDER FACE validation set, and 65.39\% in 4K-Face, which outperforms RetinaNet (65\% and 53.34\%) and UnitBox (67.8\% and 63.82\%) by $\sim$10\% in average. This results demonstrate good complementary between the anchor-based method and anchor-free method.

\subsubsection{Efficiency of Merging Strategy}
We now turn to analyze the efficiency of the proposed merging strategy. The results are illustrated in Table~\ref{table:tech_path_res} and~\ref{table:tech_path_4k}. As describe in Section~\ref{sec:classification with IOU}, directly merging the two branches leads to a significantly drop of the detection performance. With the proposed re-score method, the detection AP is boosted from 73.8\% to 80.7\% on the WIDER FACE dataset, and from 61.60\% to 65.39\% in 4K-Face.

\begin{table*}[t]
\begin{center}
\resizebox{\linewidth}{!}{
\begin{tabular}{c|cccc|cccc}
\hline\hline
BaseNet  & P3-P5 layer & re-score & Anchor-based Branch & Anchor-free Branch & AP (easy) & AP (medium) & AP (hard) \\
\hline\hline
RetinaNet & & & & & \textbf{92.6} & \textbf{91.2} & 65.0 \\
RetinaNet(multi-scale) & & & & & 90.7 & 90.3 & 75.2 \\
RetinaNet & \checkmark & & & & 43.8 & 64.9 & 74.7 \\
UnitBox & & & & & 70.6 & 76.0 & 67.8 \\
\hline
SFace & \checkmark & & \checkmark & & 43.5 & 64.4 & 73.7 \\
SFace & \checkmark & & & \checkmark & 71.6 & 78.1 & 73.7 \\
SFace & \checkmark & & \checkmark & \checkmark & 71.6 & 78.1 & 73.8 \\
SFace & \checkmark & \checkmark & \checkmark & & 39.5 & 62.4 & 72.9 \\
SFace & \checkmark & \checkmark & & \checkmark & 90.0 & 88.8 & 78.8 \\
SFace & \checkmark & \checkmark & \checkmark & \checkmark & 89.8 & 88.7 & \textbf{80.7} \\
\hline
\end{tabular}
}
\end{center}
\caption{The ablation study of SFace on the WIDER FACE validation dataset.}
\label{table:tech_path_res}
\end{table*}

\begin{table*}[t]
\begin{center}
\resizebox{\linewidth}{!}{
\begin{tabular}{c|cccc|ccccc}
\hline\hline
BaseNet & P3-P5 layer & re-score & Anchor-based Branch & Anchor-free Branch & AP($<32$) & AP($32-256$) & AP($256-512$) & AP($>1024$) & AP(all)\\
\hline\hline
RetinaNet & & & & & 1.22 & 54.62 & 74.51 & 47.74 & 53.34 \\
RetinaNet & \checkmark  & & & & \textbf{29.23} & 49.72 & 0.00 & 0.00 & 32.73 \\
UnitBox & & & & & 3.27 & 61.09 & 81 & 50.97 & 63.82 \\
\hline
SFace & \checkmark & & \checkmark & & 26.86 & 46.51 & 0.00 & 0.00 & 30.72 \\
SFace & \checkmark & & & \checkmark & 3.41 & 57.35 & 77.80 & 53.82 & 61.49 \\
SFace & \checkmark & & \checkmark & \checkmark & 3.51 & 57.38 & 77.81 & 53.80 & 61.60 \\
SFace & \checkmark & \checkmark & \checkmark & & 23.62 & 48.03 & 0.00 & 0.00 & 31.50 \\
SFace & \checkmark & \checkmark & & \checkmark & 4.07 & 62.93 & \textbf{81.38} & \textbf{60.48} & 64.30 \\
SFace & \checkmark & \checkmark & \checkmark & \checkmark & 6.70 & \textbf{63.09} & \textbf{81.35} & \textbf{60.48} & \textbf{65.39} \\
\hline
\end{tabular}
}
\end{center}
\caption{The ablation study of SFace on the 4K-Face dataset.}
\label{table:tech_path_4k}
\end{table*}

\subsubsection{Inference Time}
The speed evaluation results are illustrated in Table~\ref{table:time}. The speed are measured on NVIDIA Titan Xp GPU. By leveraging the light weight backbone network, our method demonstrates considerable run-time speed of $\sim$80fps for 1080p images, and achieves real time performance for 4K images. These results meet the practical requirements of face detection in high resolution images and videos.

\begin{table}[H]
\begin{center}
\resizebox{0.9\textwidth}{!}{
\begin{tabular}{c|c|c|c|c}
\hline
\hline
Min size & 1080 & 1200 & 1500 & 2160 \\
\hline
\hline
Time & 12.46ms & 14.30ms & 21.53ms & 41.13ms \\
AP (WIDER FACE hard) & 76.7 & 78.4 & 80.7 & 78.8 \\
\hline
\end{tabular}
}
\end{center}
\caption{The inference time and precision with respect to different input sizes for our Method.}
\label{table:time}
\end{table}

\section{Conclusion}
We present a clean and simple architecture, called SFace, to detect faces with large scale variations. The architecture integrates the anchor-based method and the anchor-free method efficiently with a novel re-score algorithm. We also present a new benchmark, 4K-Face, to better evaluate detectors in high resolution images with extremely large scale variation of faces. The proposed method demonstrates significant improvements over several strong baselines while keeping high inference speed. Thus, it provides a practical solution for research and applications of face detection, especially for high-resolution images.

\clearpage
\begin{figure*}[t]
\begin{center}
    \includegraphics[width=0.9\linewidth]{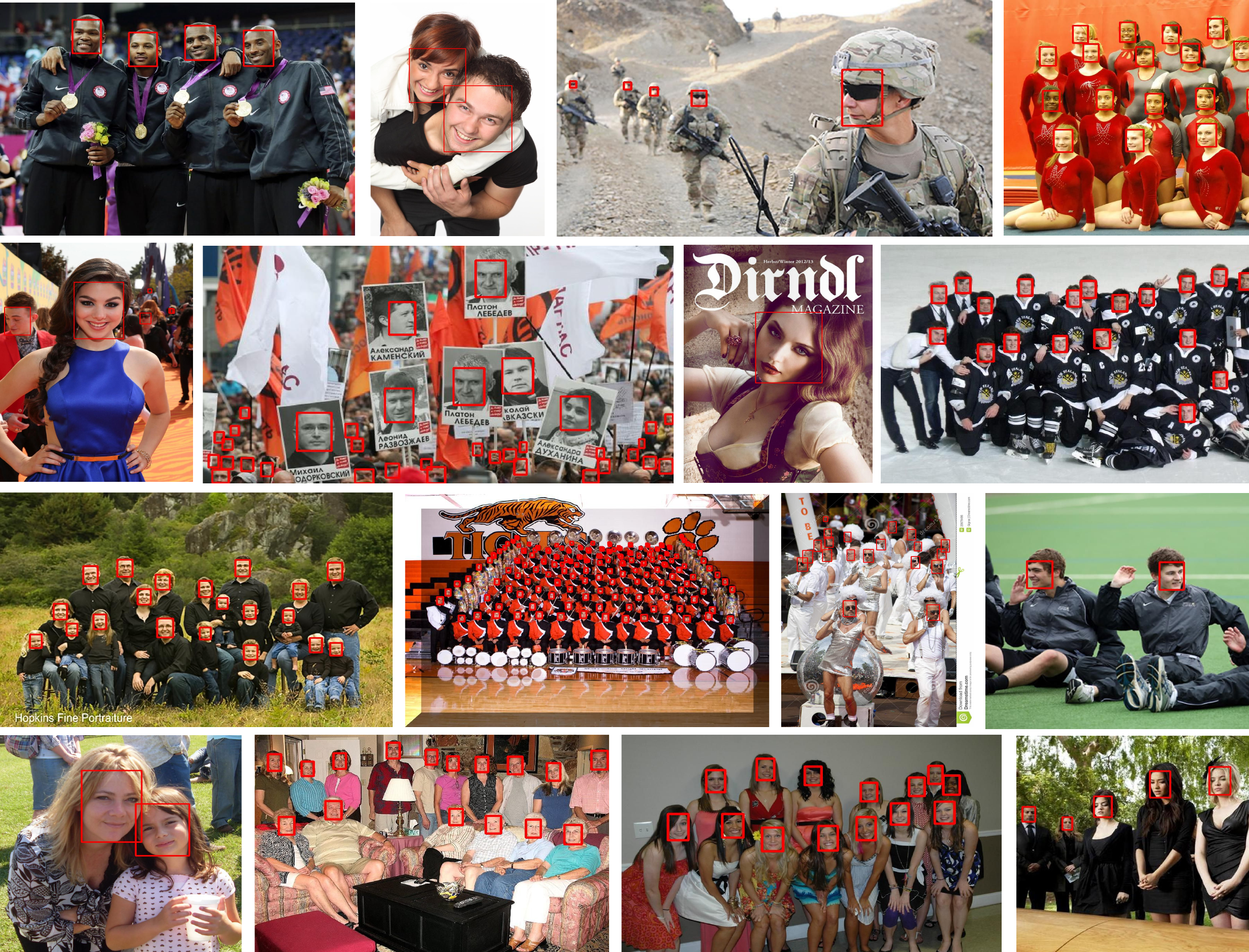}
\end{center} 
    \caption{Example results of SFace on the validation set of the WIDER FACE dataset.}
\label{fig:WIDER FACE_}
\end{figure*}

\clearpage


\bibliographystyle{splncs}
\bibliography{egbib}

\begin{thebibliography}{10}

\bibitem{huang2015densebox}
Huang, L., Yang, Y., Deng, Y., Yu, Y.:
\newblock Densebox: Unifying landmark localization with end to end object
  detection.
\newblock arXiv preprint arXiv:1509.04874 (2015)

\bibitem{yu2016unitbox}
Yu, J., Jiang, Y., Wang, Z., Cao, Z., Huang, T.:
\newblock Unitbox: An advanced object detection network.
\newblock In: Proceedings of the 2016 ACM on Multimedia Conference. (2016)
  516--520

\bibitem{lin2017focal}
Lin, T.Y., Goyal, P., Girshick, R., He, K., Dollar, P.:
\newblock Focal loss for dense object detection.
\newblock In: Proceedings of the IEEE Conference on Computer Vision and Pattern
  Recognition. (2017)  2980--2988

\bibitem{yang2016wider}
Yang, S., Luo, P., Loy, C.C., Tang, X.:
\newblock Wider face: A face detection benchmark.
\newblock In: Proceedings of the IEEE Conference on Computer Vision and Pattern
  Recognition. (2016)  5525--5533

\bibitem{xiong2013supervised}
Xiong, X., De~la Torre, F.:
\newblock Supervised descent method and its applications to face alignment.
\newblock In: Proceedings of the IEEE conference on computer vision and pattern
  recognition. (2013)  532--539

\bibitem{zhu2016face}
Zhu, X., Lei, Z., Liu, X., Shi, H., Li, S.Z.:
\newblock Face alignment across large poses: A 3d solution.
\newblock In: Proceedings of the IEEE Conference on Computer Vision and Pattern
  Recognition. (2016)  146--155

\bibitem{parkhi2015deep}
Parkhi, O.M., Vedaldi, A., Zisserman, A.,  et~al.:
\newblock Deep face recognition.
\newblock In: BMVC. Volume~1. (2015) ~6

\bibitem{schroff2015facenet}
Schroff, F., Kalenichenko, D., Philbin, J.:
\newblock Facenet: A unified embedding for face recognition and clustering.
\newblock In: Proceedings of the IEEE Conference on Computer Vision and Pattern
  Recognition. (2015)  815--823

\bibitem{zhu2015high}
Zhu, X., Lei, Z., Yan, J., Yi, D., Li, S.Z.:
\newblock High-fidelity pose and expression normalization for face recognition
  in the wild.
\newblock In: Proceedings of the IEEE Conference on Computer Vision and Pattern
  Recognition. (2015)  787--796

\bibitem{viola2001rapid}
Viola, P., Jones, M.:
\newblock Rapid object detection using a boosted cascade of simple features.
\newblock In: Computer Vision and Pattern Recognition, 2001. CVPR 2001.
  Proceedings of the 2001 IEEE Computer Society Conference on. Volume~1., IEEE
  (2001)  I--I

\bibitem{brubaker2008design}
Brubaker, S.C., Wu, J., Sun, J., Mullin, M.D., Rehg, J.M.:
\newblock On the design of cascades of boosted ensembles for face detection.
\newblock International Journal of Computer Vision \textbf{77}(1-3) (2008)
  65--86

\bibitem{zhu2006fast}
Zhu, Q., Yeh, M.C., Cheng, K.T., Avidan, S.:
\newblock Fast human detection using a cascade of histograms of oriented
  gradients.
\newblock In: Computer Vision and Pattern Recognition, 2006 IEEE Computer
  Society Conference on. Volume~2., IEEE (2006)  1491--1498

\bibitem{liao2016fast}
Liao, S., Jain, A.K., Li, S.Z.:
\newblock A fast and accurate unconstrained face detector.
\newblock IEEE transactions on pattern analysis and machine intelligence
  \textbf{38}(2) (2016)  211--223

\bibitem{felzenszwalb2008discriminatively}
Felzenszwalb, P., McAllester, D., Ramanan, D.:
\newblock A discriminatively trained, multiscale, deformable part model.
\newblock In: Proceedings of the IEEE Conference on Computer Vision and Pattern
  Recognition. (2008)  1--8

\bibitem{ren2015faster}
Ren, S., He, K., Girshick, R., Sun, J.:
\newblock Faster r-cnn: Towards real-time object detection with region proposal
  networks.
\newblock In: Advances in neural information processing systems. (2015)  91--99

\bibitem{liu2016ssd}
Liu, W., Anguelov, D., Erhan, D., Szegedy, C., Reed, S., Fu, C.Y., Berg, A.C.:
\newblock Ssd: Single shot multibox detector.
\newblock In: European conference on computer vision, Springer (2016)  21--37

\bibitem{fu2017dssd}
Fu, C.Y., Liu, W., Ranga, A., Tyagi, A., Berg, A.C.:
\newblock Dssd: Deconvolutional single shot detector.
\newblock arXiv preprint arXiv:1701.06659 (2017)

\bibitem{lin2017feature}
Lin, T.Y., Dollar, P., Girshick, R., He, K., Hariharan, B., Belongie, S.:
\newblock Feature pyramid networks for object detection.
\newblock In: Proceedings of the IEEE Conference on Computer Vision and Pattern
  Recognition. (2017)  2117--2125

\bibitem{redmon2016you}
Redmon, J., Divvala, S., Girshick, R., Farhadi, A.:
\newblock You only look once: Unified, real-time object detection.
\newblock In: Proceedings of the IEEE Conference on Computer Vision and Pattern
  Recognition. (2016)  779--788

\bibitem{redmon2017yolo9000}
Redmon, J., Farhadi, A.:
\newblock Yolo9000: Better, faster, stronger.
\newblock In: Computer Vision and Pattern Recognition (CVPR), 2017 IEEE
  Conference on, IEEE (2017)  6517--6525

\bibitem{zhang2017s3fd}
Zhang, S., Zhu, X., Lei, Z., Shi, H., Wang, X., Li, S.Z.:
\newblock S3fd: Single shot scale-invariant face detector.
\newblock In: Proceedings of the IEEE Conference on Computer Vision and Pattern
  Recognition. (2017)  192--201

\bibitem{najibi2017ssh}
Najibi, M., Samangouei, P., Chellappa, R., Davis, L.S.:
\newblock Ssh: Single stage headless face detector.
\newblock In: Proceedings of the IEEE Conference on Computer Vision and Pattern
  Recognition. (2017)  4875--4884

\bibitem{zhang2016joint}
Zhang, K., Zhang, Z., Li, Z., Qiao, Y.:
\newblock Joint face detection and alignment using multitask cascaded
  convolutional networks.
\newblock IEEE Signal Processing Letters \textbf{23}(10) (2016)  1499--1503

\bibitem{chollet2017xception}
Chollet, F.:
\newblock Xception: Deep learning with depthwise separable convolutions.
\newblock In: Proceedings of the IEEE Conference on Computer Vision and Pattern
  Recognition. (2017)  1251--1258

\bibitem{russakovsky2015imagenet}
Russakovsky, O., Deng, J., Su, H., Krause, J., Satheesh, S., Ma, S., Huang, Z.,
  Karpathy, A., Khosla, A., Bernstein, M.,  et~al.:
\newblock Imagenet large scale visual recognition challenge.
\newblock International Journal of Computer Vision \textbf{115}(3) (2015)
  211--252

\bibitem{hao2017scale}
Hao, Z., Liu, Y., Qin, H., Yan, J., Li, X., Hu, X.:
\newblock Scale-aware face detection.
\newblock In: Proceedings of the IEEE Conference on Computer Vision and Pattern
  Recognition. (2017)  6186--6195

\bibitem{girshick2015fast}
Girshick, R.:
\newblock Fast r-cnn.
\newblock In: Proceedings of the IEEE international conference on computer
  vision. (2015)  1440--1448

\bibitem{loshchilov2017fixing}
Loshchilov, I., Hutter, F.:
\newblock Fixing weight decay regularization in adam.
\newblock arXiv preprint arXiv:1711.05101 (2017)

\bibitem{howard2013some}
Howard, A.G.:
\newblock Some improvements on deep convolutional neural network based image
  classification.
\newblock arXiv preprint arXiv:1312.5402 (2013)

\bibitem{neubeck2006efficient}
Neubeck, A., Van~Gool, L.:
\newblock Efficient non-maximum suppression.
\newblock In: Pattern Recognition, 2006. ICPR 2006. 18th International
  Conference on. Volume~3., IEEE (2006)  850--855

\bibitem{yang2014aggregate}
Yang, B., Yan, J., Lei, Z., Li, S.Z.:
\newblock Aggregate channel features for multi-view face detection.
\newblock In: Biometrics (IJCB), 2014 IEEE International Joint Conference on,
  IEEE (2014)  1--8

\bibitem{yang2015facial}
Yang, S., Luo, P., Loy, C.C., Tang, X.:
\newblock From facial parts responses to face detection: A deep learning
  approach.
\newblock In: Proceedings of the IEEE International Conference on Computer
  Vision. (2015)  3676--3684

\bibitem{ohn2016boost}
Ohn-Bar, E., Trivedi, M.M.:
\newblock To boost or not to boost? on the limits of boosted trees for object
  detection.
\newblock In: Pattern Recognition (ICPR), 2016 23rd International Conference
  on, IEEE (2016)  3350--3355

\bibitem{zhu2017cms}
Zhu, C., Zheng, Y., Luu, K., Savvides, M.:
\newblock Cms-rcnn: contextual multi-scale region-based cnn for unconstrained
  face detection.
\newblock In: Deep Learning for Biometrics.
\newblock Springer (2017)  57--79

\bibitem{yang2017face}
Yang, S., Xiong, Y., Loy, C.C., Tang, X.:
\newblock Face detection through scale-friendly deep convolutional networks.
\newblock arXiv preprint arXiv:1706.02863 (2017)

\bibitem{Hu_2017_CVPR}
Hu, P., Ramanan, D.:
\newblock Finding tiny faces.
\newblock In: The IEEE Conference on Computer Vision and Pattern Recognition
  (CVPR). (July 2017)

\bibitem{he2016deep}
He, K., Zhang, X., Ren, S., Sun, J.:
\newblock Deep residual learning for image recognition.
\newblock In: Proceedings of the IEEE conference on computer vision and pattern
  recognition. (2016)  770--778

\end{thebibliography}
\end{document}